\documentclass[a4paper,fleqn]{cas-dc}

\usepackage[numbers]{natbib}

\begin{document}
\let\WriteBookmarks\relax
\def\floatpagepagefraction{1}
\def\textpagefraction{.001}

\title[mode = title]{Implementing blind navigation through multi-modal sensing and gait guidance}
\tnotemark[1]

\tnotetext[1]{This document is the results of the research project
	funded by the National Science Foundation.}

\author[1]{Feifan Yan}[style=chinese]
\cormark[1] 
\author[1]{Tianle Zeng}[style=chinese]
\author[1]{Meixi He}[style=chinese]
\address[1]{China University of Mining and Technology - Beijing}
\shortauthors{F. Yan, et al.}


\cortext[1]{Corresponding author:***}

\begin{abstract}[S U M M A R Y]
By the year 2023, the global population of individuals with impaired vision has surpassed 220 million. People with impaired vision will find it difficult while finding path or avoiding obstacles, and must ask for auxiliary tools for help. Although traditional aids such as guide canes and guide dogs exist, they still have some shortcomings. In this paper, we present our wearable blind guiding device, what perform navigation guidance through our proposed Gait-based Guiding System. Our device innovatively integrates gait phase analysis for walking guide, and in terms of environmental perception, we use multimodal sensing to acquire diverse environment information. During the experiment, we conducted both indoor and outdoor experiments, and compared with the standard guide cane. The result shows superior performance of our device in blind guidance. 
\end{abstract}

\begin{highlights}
	\item Integration of multimodal sensors enhances navigation capability in various settings.
	
	\item Novel method provides effective guidance for the visually impaired.
	
	\item All wearable device enhances mobility capability of vision impared.
\end{highlights}

\maketitle

\section{Inroduction}\label{Indro}
As of 2023, the global population of individuals with impaired vision exceeds 220 million \cite{world2019world}. The absence of visual perception among these individuals gives rise to both physical and psychological challenges, manifesting in issues such as reduced efficiency in daily travel, difficulty navigating unfamiliar environments, and an inability to navigate obstacles accurately \cite{crewe2011quality}. Coping with the extended treatment cycles common in vision-related ailments necessitates the use of aids to navigate the period of visual impairment \cite{bourne2017magnitude}. Traditional guidance tools like guide dogs and canes, while commonly employed, fall short in providing accurate guidance in intricate and unfamiliar surroundings. Moreover, the use of these aids poses a challenge by occupying the hands of individuals with visual impairments, causing interference in their daily lives. The training duration and costs associated with guide dogs, along with the prolonged adaptation period for guide canes, further compound the challenges faced by individuals with visual impairments.

There are two main challenges in blind guidance: obstacle avoidance, indoor and outdoor guidance. Numerous blind guidance devices address these challenges by incorporating diverse sensors for environment perception. Decision algorithms play a pivotal role in path planning while guiding the visually impaired. Auxiliary blind guiding devices commonly leverage GPS technology \cite{lakde2015review} for outdoor navigation. GIS \cite{fernandes2012providing} is a powerful tool for navigation, providing geographic and environmental information. In indoor occasion, Light Detection and Ranging (LIDAR) \cite{mai2023laser} usually integrated into some devices for expansive distance sensing, Simultaneous Localization and Mapping (SLAM) \cite{taheri2021slam} being a powerful algorithm in this occasion, for map construction and positioning with a LIDAR \cite{khan2021comparative}, camera \cite{macario2022comprehensive} or millimeter wave radar \cite{li2020millimeter}. Furthermore, computer vision proves to be an effective solution for both object identification and obstacle avoidance \cite{al2016ebsar}. 

Challenges also exist in the approach to guiding the blind along specific paths. The approach should be robust, while not interfering the daily mobility of blind. It is imperative that any solution is fully wearable and imposes no additional cognitive load. Regular methods already exist, such as tactile \cite{lee2014wearable}, auditory \cite{altaha2016blindness}, guiding dogs, force from grounded kinesthetic assist \cite{slade2021multimodal}. Problems still exists, as these approaches fall short in fully meeting the needs of the visually impaired. 

Recognizing the limitations of conventional methods, our approach stems from the segmentation of human periodic walking into gait phases based on the motion of the lower limbs \cite{sethi2022comprehensive}.  Derived from clinical medical analysis, gait cycle phase is defined as a rhythmic movement of the human lower limbs during walking, what makes human body walking forward \cite{bensoussan2008evaluation}. Gait analysis, a systematic method for recognizing human walk \cite{whittle2014gait,kirtley2006clinical,gage1995gait}, has found applications in clinical diagnosis, identity recognition, and robotics \cite{coutts1999gait,jacquelin2010gait,qiu2019body,shih2012stable,kale2004identification,veneman2007design}. In the realm of wearable assistive devices, successful collaboration between humans and machines is essential for providing seamless and harmonious assistance \cite{tao2012gait}. This underscores the rationale behind incorporating gait phases into our innovative device. 

\begin{figure*}
	\centering
		\includegraphics[scale=.35]{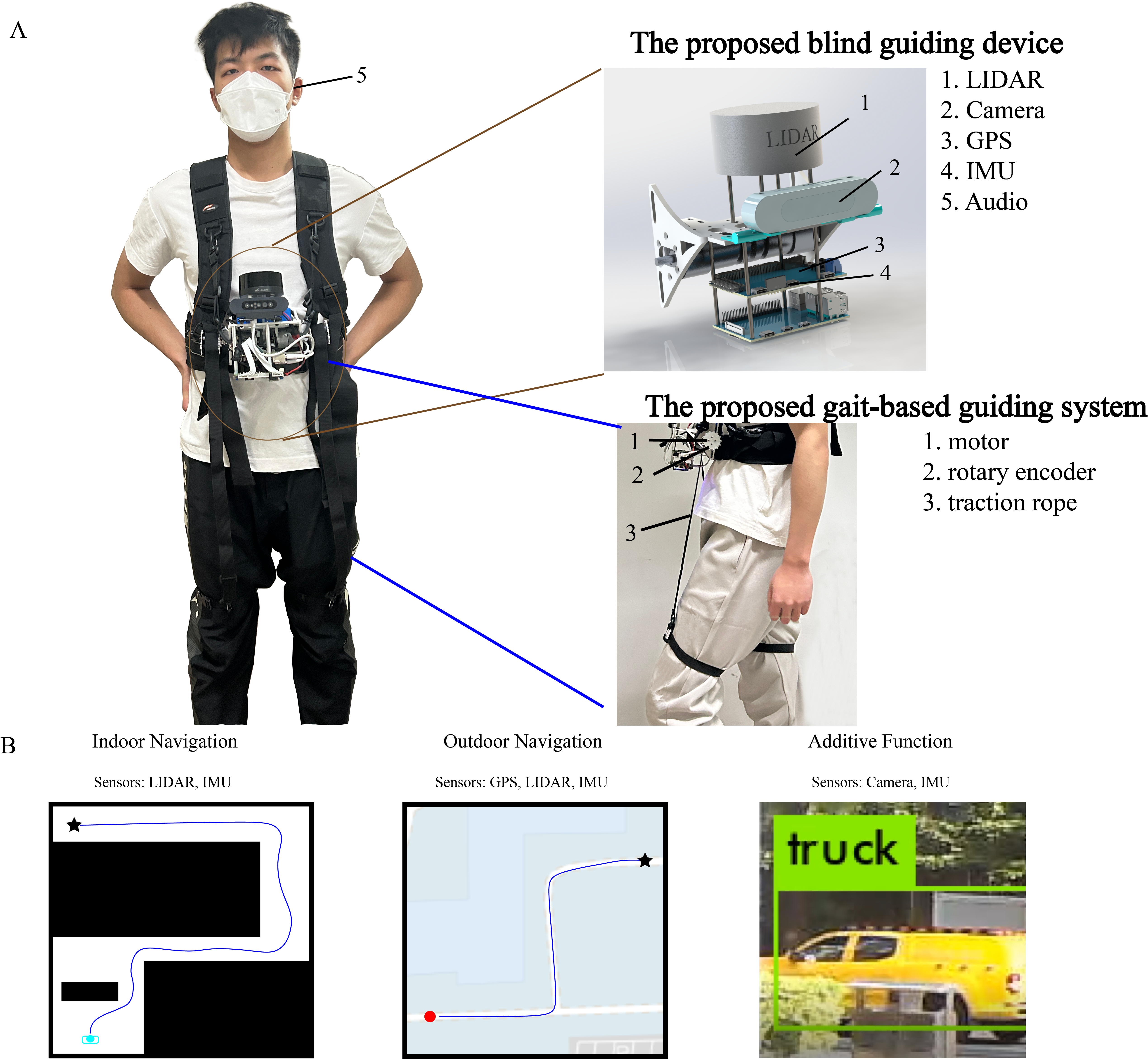}
	\caption{Overview of the Proposed Blind Guiding Device. (A) The device incorporates multiple sensors and our innovative gait-based guiding system. (B) Various sensors are employed to capture environmental information across different scenarios. }
	\label{FIG:1}
\end{figure*}

Then it comes to our proposed wearable blind guiding device, depicted in Fig. \protect\ref{FIG:1}(A). The device integrates 2D LIDAR, GPS, Camera, and inertial measurement unit (IMU), to provide environmental conception. Also integrates the proposed Gait-based Guiding System, which consists of two motors and two traction ropes tied to the thigh, the system could be able to provide guidance to specific paths through sensing and affecting gait phase of blind while walking in appropriate gait phase. The total weight of the equipment is 2kg, and the research and development cost amounted to approximately \$400, mass production could be as low as around \$200. 

The structure of the remainder of this paper is as follows: Section 2 outlines the methodology of our proposed device, including the details of our innovative gait-based guiding system. In Section 3, we present the results of our device, delineate the experimental procedures, and highlight the performance of our method in comparison to traditional approaches. Section 4 delves into detailed discussions. Lastly, Section 5 provides conclusive remarks.

\section{Methodology}
In the realm of blind guiding devices, two essential functions must be addressed: (a) environmental perception and route planning, and (b) guiding individuals with visual impairments along specific paths. 

The environmental perception aspect is systematically addressed through the strategic utilization of multimodal sensors, with a more detailed exposition provided in subsequent sections. 

The proposed method: Gait-based Guiding System is called out to surmount the challenge (b), what could guide human walking direction in proper gait phases. Analysis of gait phase will be carried out, as well as our methodology of gait cycle recognition and gait guidance while walking, to provide a nuanced solution. 

\subsection{Environment perception}
The proposed device is designed to cater to the demands of daily use, prioritizing lightweight design and portability while ensuring minimal environmental impact. A suite of multiple sensors is incorporated to compensate for the visual information lost by individuals with visual impairments, facilitating efficient navigation toward predefined goals (see Fig. \protect\ref{FIG:1}(B)). 

Global Positioning System (GPS) stands out as the most effective navigation method, boasting an accuracy of less than 2 meters in global positioning. To enhance outdoor navigation precision, our novel Gait-based Guiding System doubles as a pedometer when coupled with GPS. Additionally, LIDAR provides effective distance perception within a 30-meter range, playing a pivotal role in obstacle avoidance. It employs a straightforward algorithm, delivering commendable results in both indoor and outdoor settings. Some methods are used to enhance the effectiveness of our instruments \cite{zeng2024realistic}\cite{zeng2024yoco}.

In settings where GPS signals are inaccessible, particularly indoors, Simultaneous Localization and Mapping (SLAM) is employed with 2D LIDAR. This technology aids in positioning and map construction, complemented by a pathfinding algorithm to proficiently navigate indoor spaces and avoid obstacles. The integration of a camera serves as an additive feature, employing the YOLOv3-tiny model for object recognition across 80 categories. Audio cues inform users upon the recognition of objects of interest. 

For the device's structural integrity, universal M3 screws and plastic studs are utilized, offering accessibility and ease of assembly. Employing an additive manufacturing process, specifically 3D printing, ensures a cohesive design that interconnects each module seamlessly. This approach minimizes material waste and contributes to a lighter overall weight. The use of universally available components, coupled with the simplicity of the assembly process—free from welding or other intricate operations—renders the device easily procurable and user-friendly. Additionally, strategically placed M3 screw holes are reserved for the potential incorporation of additional sensors in the future.

\begin{figure*}
	\centering
		\includegraphics[scale=.45]{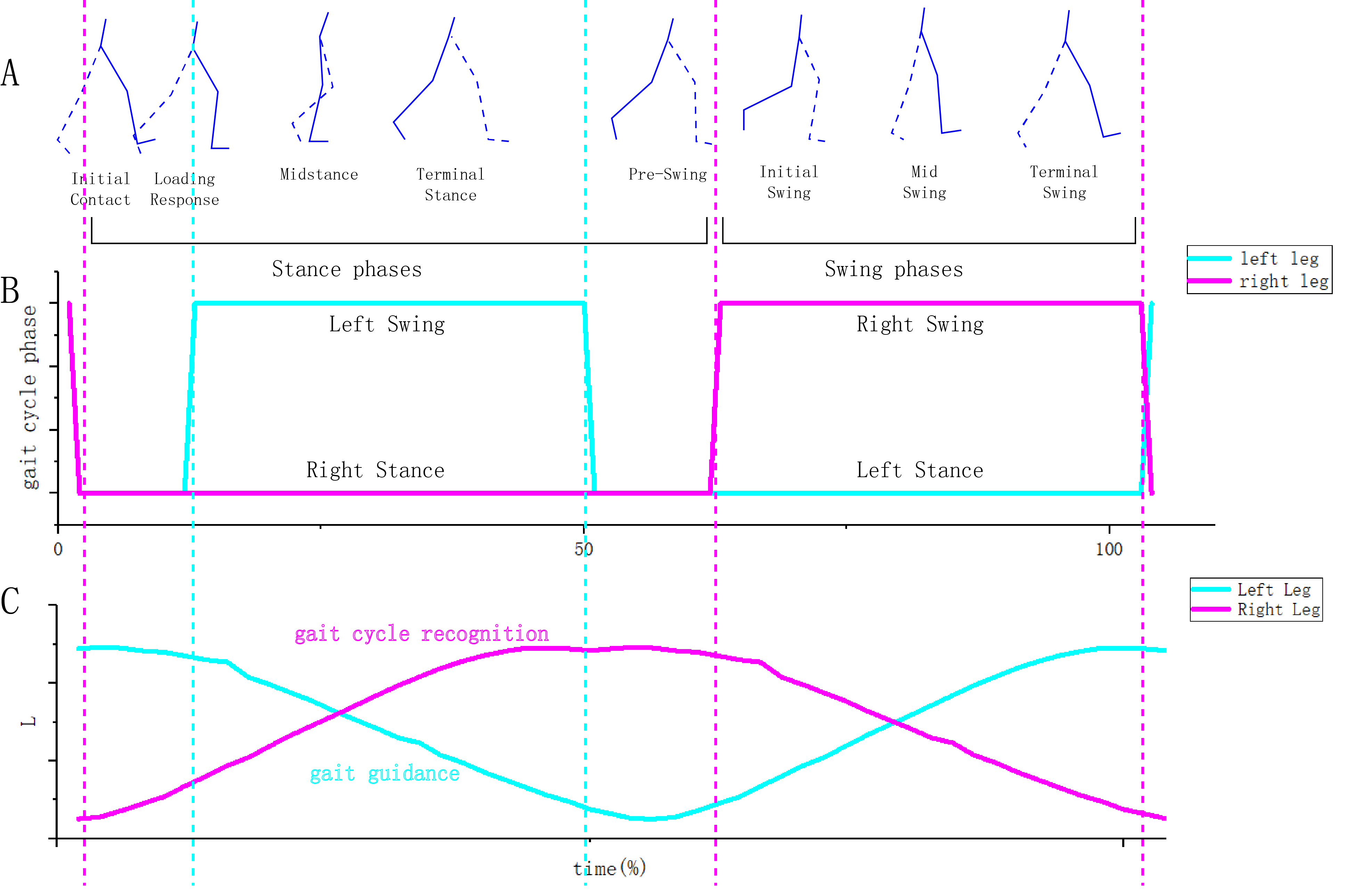}
	\caption{Gait Cycle Phases. (A) Human lower limb gait cycle phases, comprising 2 main phases: stance phase and swing phase, further segmented into 8 sub-phases. (B) The temporal distribution of each gait cycle phase. (C) The time curve of $\mathcal{L}$, illustrating the periods of gait cycle recognition and guidance on the right leg.}
	\label{FIG:2}
\end{figure*}

\subsection{Gait-based Guiding System}
Human walk is inherently a periodic process, characterized by distinct swing and stance phases that can be further segmented into various stages \cite{drnach2018identifying}. As illustrated in Fig. \protect\ref{FIG:2}(a), the phases of gait have been divided into 8 phases based on foot movement: initial contact, loading response, midstance, terminal stance, pre-swing, initial swing, mid swing, and terminal swing \cite{gage1995gait}. The temporal progression of these phases is depicted in the time curve presented in Fig. \protect\ref{FIG:2}(b). Various approaches for gait analysis have been employed, as documented in existing literature \cite{liu2016gait,gurchiek2019open}. Recognizing the significance of the thigh in the walking process \cite{ozaki2011increases,lovejoy2005natural}, our focus is on utilizing thigh motion to regulate the stride length of each leg. 

Fig. \protect\ref{FIG:3}(a) showcases the human thigh model alongside our proposed Gait-based Guiding System. This model consists of the human limb and our guiding system, capable of sensing lower limb movements during walking and accurately recognizing different phases of the gait cycle.

\subsubsection{gait phase recognition and guidance}
\begin{figure}
	\centering
		\includegraphics[scale=.45]{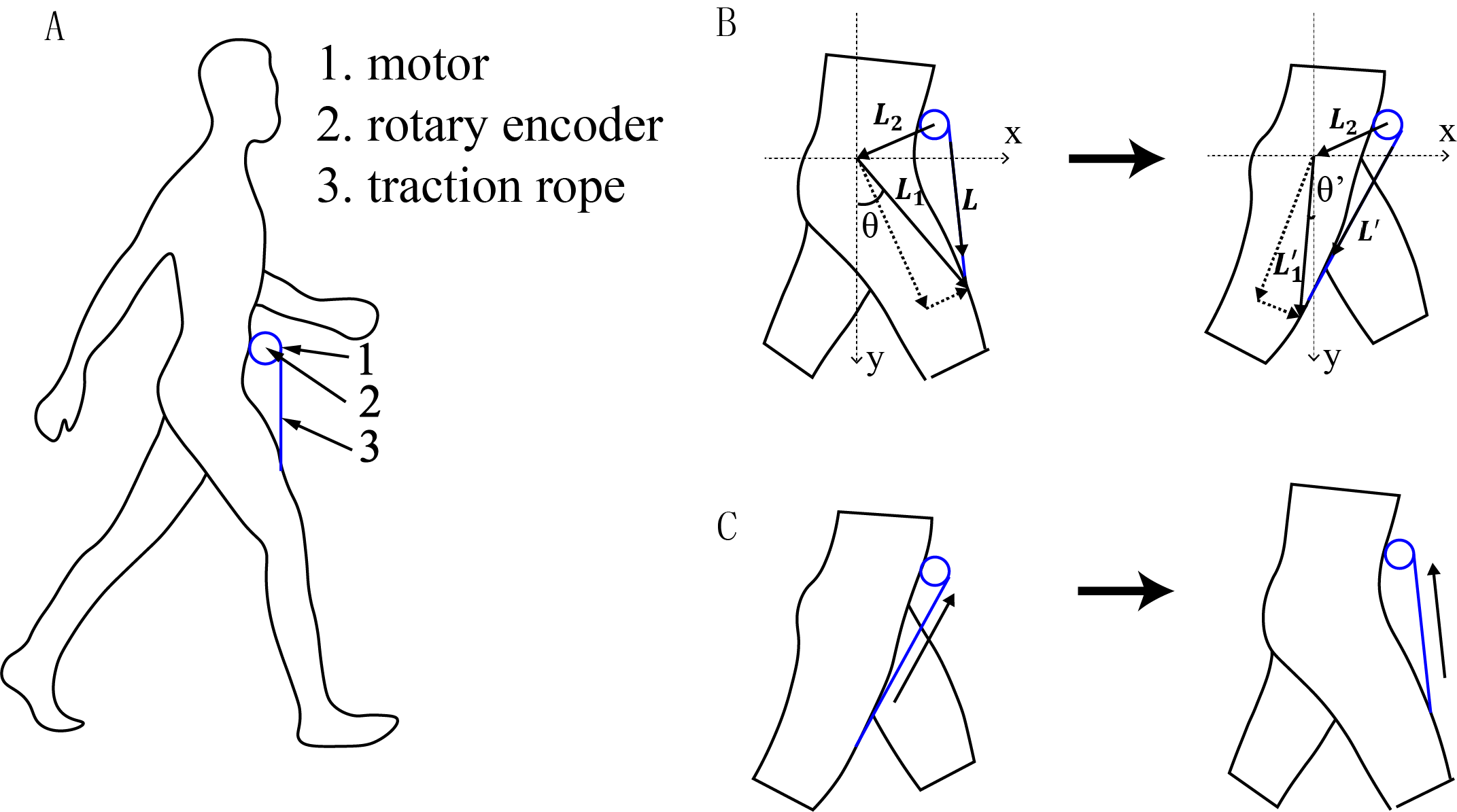}
	\caption{Overview of the Proposed Gait-Based Guiding System. (A) The Gait-Based Guiding System comprises two motors positioned in front of the human stomach, equipped with rotary encoders, and two traction ropes connecting the motors to the human thighs. (B) The alteration in traction rope length (L) serves as a means for sensing the gait cycle. (C) Auxiliary force from the motors, applied at specific intervals, facilitates gait guidance, addressing asymmetry between the inner and outer legs.}
	\label{FIG:3}
\end{figure}

Recognition of the human walking gait cycle is facilitated through the incorporation of rotary encoders mounted on the motors. Specifically, during the stance phase (refer to Fig. \protect\ref{FIG:2}(C)), thigh movement is measured through these rotary encoders by sensing changes in the length (L) of the traction ropes, as illustrated in Fig. \protect\ref{FIG:3}(b). By the way, a controlled amount of damping provided by the motors being needed, so that the traction ropes can fully adapt to motion of thigh while does not affect normal mobility. 
In general terms, the human thigh can be considered as a rigid body. The dynamic equation describing the movement of two legs is: 
\[L'_1=R(\theta)*L_1\]
  \[L=L_1+L_2\]
then:
  \[L'=L'_1+L_2=R(\theta)*L_1+L_2\]

In the provided equation, $\mathcal{L}$ and $\mathcal{L}'$ represent the vectors of the traction rope while walking, $\mathcal{L}_1$ and $\mathcal{L}'_1$ are the vectors from hip joint to the thigh while walking. Integrating these vectors with the movement of human legs during walking allows us to construct a time curve for $\mathcal{L}$, as illustrated in Fig. \protect\ref{FIG:2}(C). 

\subsubsection{Steering guidance based on gait}
Previous research has demonstrated that the steering of the human body involves asymmetry in stride length and phase differences between the movements of the inner and outer legs \cite{courtine2003humanII}. The theory has found practical application in steering strategies for certain bipedal robots \cite{shih2012stable}. As discussed in the preceding section, our approach involves gait control during specific phases to accomplish the objective of steering guidance. Introducing an asymmetry in the control strategy becomes instrumental in providing guidance for steering during walking. 

Our focus is on influencing the stride length and walking phase of each leg, achievable through the coordinated action of motors and traction ropes tethered to the thigh. This effect is further elucidated in Fig. \protect\ref{FIG:3}(C), illustrating the application of auxiliary force based on our Gait-based Guiding System during the swing phase (as depicted in Fig. \protect\ref{FIG:2}(C)). 

\begin{figure}
	\centering
		\includegraphics[scale=.57]{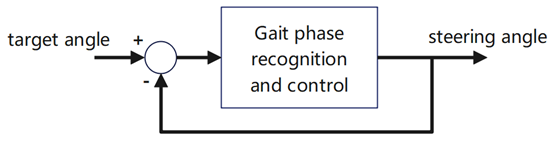}
	\caption{Block Diagram of the Proposed Gait-Based Guiding System for Steering. The steering angle is sensed by the Inertial Measurement Unit (IMU), and the steering guidance is facilitated through the Gait Phase Recognition and Guidance, as discussed in the preceding section.}
	\label{FIG:4}
\end{figure}

Illustrated in Fig. \protect\ref{FIG:4}, with the successful control of the gait phase and the incorporation of feedback through the Inertial Measurement Unit (IMU), a closed-loop steering guide based on gait can be effectively achieved. Given that the steering torque is not exclusively generated by the lower limbs \cite{courtine2003humanI,patla1999online}, audio instructions become imperative, especially during significant steering angles, to offer advance reminders. 

To conserve energy, the traction rope is designed to relax during straight walking and tighten when guidance is required. The closed-loop control of our proposed Gait-based Guiding System can seamlessly integrate automatic control algorithms, enhancing the overall efficiency and adaptability of the system. 

\section{Results}
We organized experiments on the functions of our device, such as obstacle avoidance and navigation. All participants wore eye masks, with an average age of 23 years old. Since the walking speed and walking route being the most intuitive evaluation indicator, we also conducted comparative experiments between the proposed device and white canes. 

\subsection{Gait-based Guiding System experiment}
\begin{figure*}
	\centering
		\includegraphics[scale=.90]{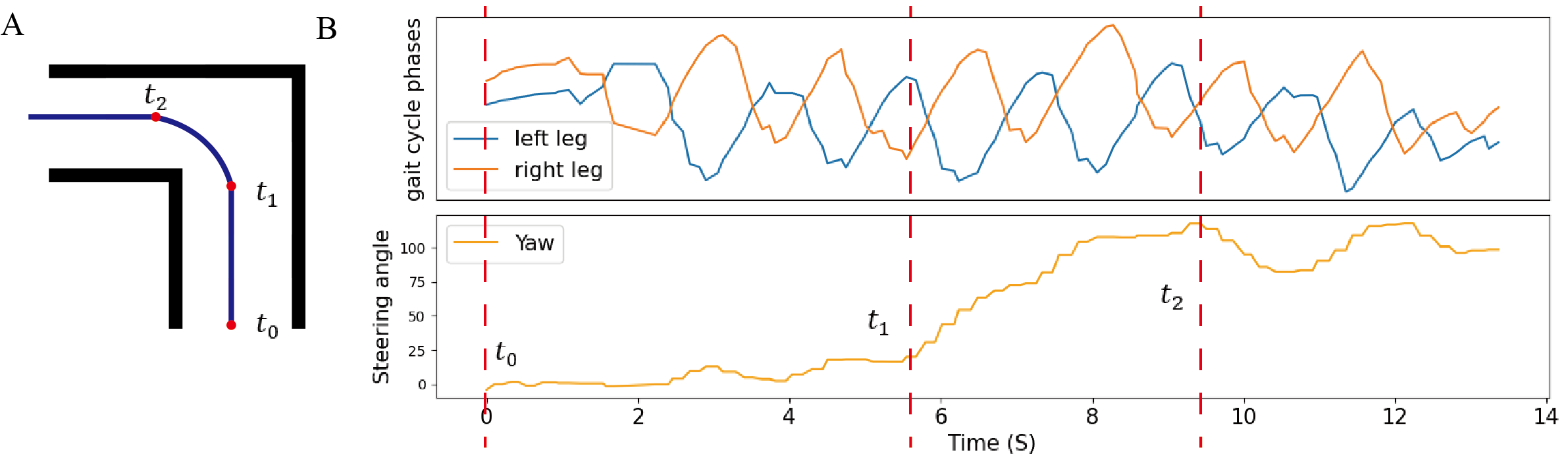}
	\caption{Gait-Based Guiding System Experiment. (A) Visualization of the experimental scenario. At time $t_1$, human gaits transition into asymmetry, and at time $t_2$, human gaits return to symmetry. (B) Time curve of $\mathcal{L}$ during the depicted path. At time $t_1$, human gaits transition into asymmetry, and at time $t_2$, human gaits return to symmetry. The time curve of the steering angle is also illustrated below.}
	\label{FIG:5}
\end{figure*}

In the experiment conducted on our proposed Gait-based Guiding System, participants followed the path outlined in Fig. \protect\ref{FIG:5}(A). The time curve of $\mathcal{L}$ and the response curve of the steering angle during detected steering are presented in Fig. \protect\ref{FIG:5}(B), asymmetric on left leg during steering being significant in the time curve of the right leg, while the left leg maintained relative symmetry in contrast. During a left turn, where the right leg functions as the outer leg, gait guidance was applied to the right leg during swing phase. This led to the asymmetry in both time and stride length. 

Across multiple experiments (n=10), where volunteers' stride length was approximately 0.45m and walking speed was around 0.8m/s, our proposed method consistently achieved a 90° curve path crossing within 2.5 seconds. 

\begin{figure}
	\centering
		\includegraphics[scale=.55]{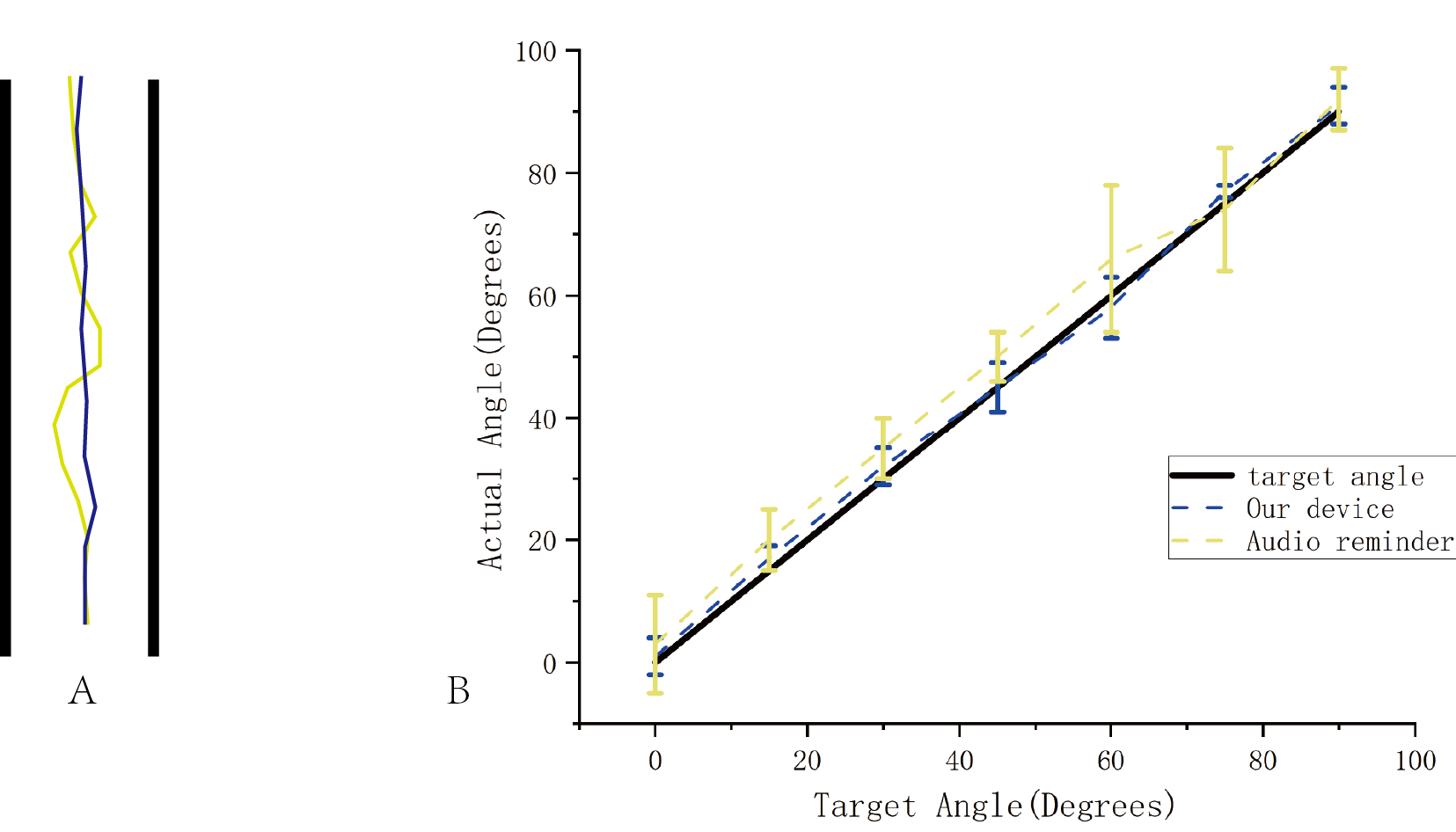}
	\caption{Comparative Experiments between Individuals Wearing Our Device and those with Audio Reminder. (A) Straight walking experiment: Volunteers were instructed to walk straight in a hall for twenty seconds, and the recorded route was analyzed. (B) Steering experiment: Five volunteers were directed to steer to a specific angle, and the actual angle after ten seconds was recorded for analysis.}
	\label{FIG:6}
\end{figure}

Straight walking and steering to an accurate angle being hard for human with impaired vision. We conducted experiments to test the capability of our device in straight walking and steering guidance,illustrated in Fig. \protect\ref{FIG:6},  Comparative experiments being conducted, and recorded the route and time spent, between blinds wearing our device and with audio reminder. 

As is shown in Fig. \protect\ref{FIG:6}(A), for straight walking experiment, the result showed robust guiding ability of our device. Compared to people with audio reminder, volunteers wearing our assistive devices achieved a smoother straight route, this gap might be higher since we had not calculated the situation where people without assistance lose their direction. As for steering experiment, shown in Fig. \protect\ref{FIG:6}(B), the actual angle after ten seconds was recoded. The result showed an improvement in accuracy of steering with an RMSE of 1.6036±1.4898, compared with people with no assist of 4.2258±8.1429. The result showed a robust of our gait-based guiding system in blind guidance, a significant improvement in the capability of walking has shown in Fig. \protect\ref{FIG:5} of blind before and after wearing our device. The improvement, however, would be highly influenced by walking speed. 

\subsection{Indoor and outdoor navigation experiment}
Indoor and outdoor navigation experiments were also conducted, to analysis the mobility of visually impaired individuals while using our blind guiding device, in comparison to a white cane. 

\begin{figure*}
	\centering
		\includegraphics[scale=.50]{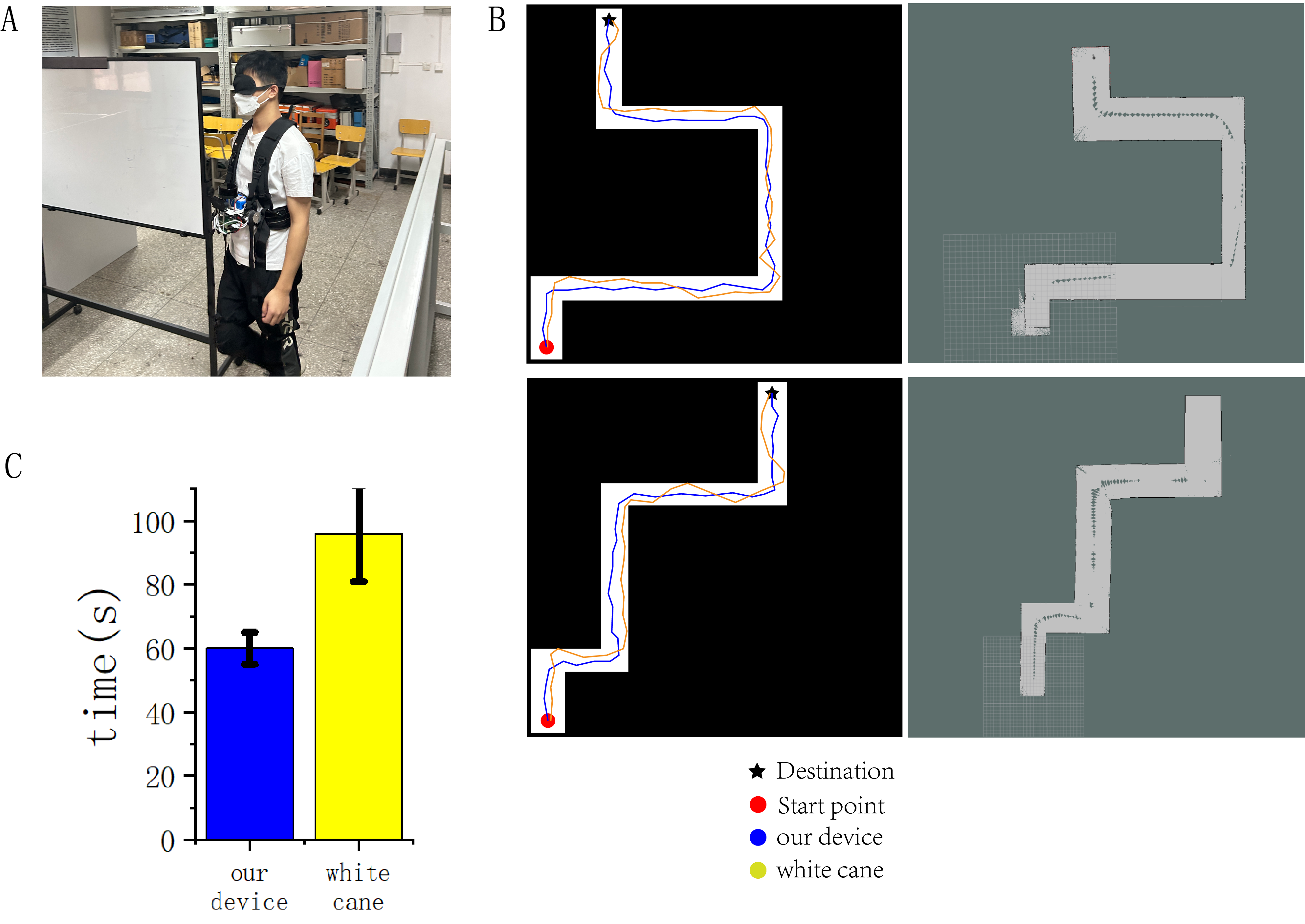}
	\caption{Indoor Navigation Experiment. (A) Obstacle avoidance experiment. (B) Experiment on Indoor Hallway Navigation. Two experimental sites are illustrated on the left side, with the walking route outlined. LIDAR-based SLAM processing in this scenario generates a mapping of the environment while localizing the user, as shown on the right side. (C) Diagram illustrating the time spent during the indoor hallway navigation experiment. }
	\label{FIG:7}
\end{figure*}

The indoor experiments comprised obstacle avoidance and an indoor hallway traversal. Five volunteers wearing our proposed device participated in these experiments. For the obstacle avoidance experiment, as illustrated in Fig. \protect\ref{FIG:7}(A), volunteers wearing our device were tasked with navigating obstacles without physical contact with the environment. The success rate was recorded and compared with the performance of a white cane. Our device demonstrated an impressive 96\% probability of successfully bypassing obstacles, outperforming the white cane, which achieved approximately 90\%. While individual results may vary, the significant improvement in obstacle avoidance capability is evident in our device. For indoor hallway experiment, in occasions shown in figure 7(B), volunteers wearing our device were asked to come across the hallway, route and time spent were recoded, compared with white cane. Our proposed device achieved Smoother route, as well as faster mobility, compared with white cane, with an speed improvement of approximately 23\%. 

\begin{figure}
	\centering
		\includegraphics[scale=.50]{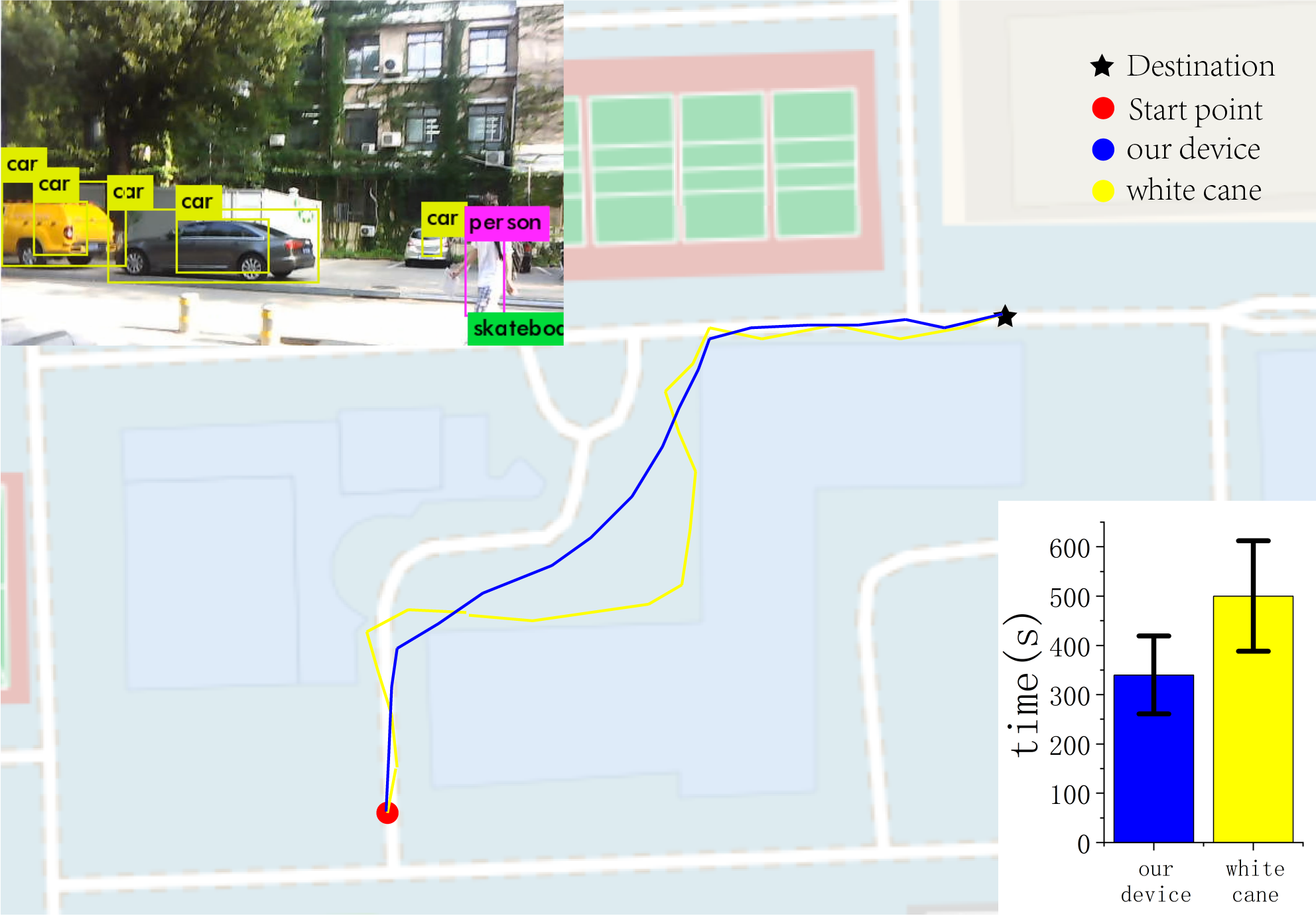}
	\caption{Outdoor Navigation Experiment. The top left corner features a demonstration of our computer vision. The bar chart with error bars in the bottom right corner illustrates the time spent by our device and a white cane.}
	\label{FIG:8}
\end{figure}

In the outdoor navigation experiment, participants were informed of the route before the experiment's commencement. From the designated starting point, volunteers, equipped with either our device or a white cane, autonomously navigated to the destination without additional assistance. As illustrated in Fig. \protect\ref{FIG:8}, our device demonstrated a smoother route with reduced physical contact with the surroundings, leading to an average time spent approximately 32.0\% less than that with the white cane. Additionally, we implemented the yoloV3-tiny model for object recognition. Detected objects are visually conveyed to the user through video, enabling them to make informed decisions to either avoid or approach the detected objects.

\section{Discussion}
This paper introduces a novel blind guiding method, the proposed wearable blind guiding device, which incorporates our novel Gait-based Guiding System and multiple sensors to offer effective guidance for individuals with impaired vision. The gait-based guiding system, composed of two motors with rotary encoders and two traction ropes tied to the thigh, demonstrates simplicity, convenience, and wearability. Our experiments underscore its effectiveness in blind guidance, showcasing superior performance compared to traditional white canes, resulting in an speed improvement of approximately 32\% compared to traditional white canes.

One significant advantage of our gait-based guidance system is its simplicity and adaptability. The system only has two motors and a traction rope, which is prone to wear and tear. With sensors and batteries, its total weight is about 2KG, which will not interfere with daily activities. Its effectiveness is evident in various experiments, especially in terms of route optimization and time efficiency compared to a white cane. It is worth noting that our system addresses the challenge of guiding visually impaired individuals by focusing on the inherent characteristics of gait, thereby allowing for broader adaptability across different users without requiring excessive training.

While our proposed approach excels in providing fundamental guidance for individuals with impaired vision, some limitations exist. The device currently lacks the capability to facilitate stair navigation, and its environmental perception is limited to a 2D LIDAR sensor, restricting omni-directional awareness. Integrating additional methods or sensors could mitigate these limitations. 

In conclusion, our proposed wearable blind guiding device, with its innovative gait-based guiding system, presents a promising solution for blind guidance. While it exhibits notable advantages in simplicity, wearability, and adaptability, addressing current limitations would further enhance its utility and broaden its applicability in diverse environments.

\section{Conclusion}
This study introduces a novel device for blind guidance, leveraging multimodal sensors to effectively compensate for the visual deficits experienced by individuals with blindness. The integration of our gait-based guiding system enables efficient navigation for the blind, incorporating gait recognition throughout the walking process and offering targeted assistance during specific gait stages to facilitate turns. Notably, our method demonstrates an overall improvement of approximately 30\% in walking speed compared to traditional white canes.

The distinctive advantages of our proposed method encompass precise steering guidance, minimal environmental contact, reduced cognitive load, and enhanced compatibility with the human body. These attributes contribute to an improved quality of life for visually impaired individuals, fostering independence and normalcy.

The conducted experiments validate the effectiveness of our device in navigation, showcasing its potential to significantly impact the lives of the visually impaired. With its cost-effectiveness and inherent advantages, our device holds promise in increasing the prospects of independent living for blind individuals.

\bibliographystyle{elsarticle-num}
\bibliography{cas-refs}

\begin{thebibliography}{10}
\expandafter\ifx\csname url\endcsname\relax
  \def\url#1{\texttt{#1}}\fi
\expandafter\ifx\csname urlprefix\endcsname\relax\def\urlprefix{URL }\fi
\expandafter\ifx\csname href\endcsname\relax
  \def\href#1#2{#2} \def\path#1{#1}\fi

\bibitem{world2019world}
W.~H. Organization, et~al., World report on vision (2019).

\bibitem{crewe2011quality}
J.~M. Crewe, N.~Morlet, W.~H. Morgan, K.~Spilsbury, A.~Mukhtar, A.~Clark, J.~Q. Ng, M.~Crowley, J.~B. Semmens, Quality of life of the most severely vision-impaired, Clinical \& experimental ophthalmology 39~(4) (2011) 336--343.

\bibitem{bourne2017magnitude}
R.~R. Bourne, S.~R. Flaxman, T.~Braithwaite, M.~V. Cicinelli, A.~Das, J.~B. Jonas, J.~Keeffe, J.~H. Kempen, J.~Leasher, H.~Limburg, et~al., Magnitude, temporal trends, and projections of the global prevalence of blindness and distance and near vision impairment: a systematic review and meta-analysis, The Lancet Global Health 5~(9) (2017) e888--e897.

\bibitem{lakde2015review}
C.~K. Lakde, P.~S. Prasad, Review paper on navigation system for visually impaired people, International Journal of Advanced Research in Computer and Communication Engineering 4~(1) (2015) 166--168.

\bibitem{fernandes2012providing}
H.~Fernandes, N.~Concei{\c{c}}{\~a}o, H.~Paredes, A.~Pereira, P.~Ara{\'u}jo, J.~Barroso, Providing accessibility to blind people using gis, Universal Access in the Information Society 11 (2012) 399--407.

\bibitem{mai2023laser}
C.~Mai, D.~Xie, L.~Zeng, Z.~Li, Z.~Li, Z.~Qiao, Y.~Qu, G.~Liu, L.~Li, Laser sensing and vision sensing smart blind cane: A review, Sensors 23~(2) (2023) 869.

\bibitem{taheri2021slam}
H.~Taheri, Z.~C. Xia, Slam; definition and evolution, Engineering Applications of Artificial Intelligence 97 (2021) 104032.

\bibitem{khan2021comparative}
M.~U. Khan, S.~A.~A. Zaidi, A.~Ishtiaq, S.~U.~R. Bukhari, S.~Samer, A.~Farman, A comparative survey of lidar-slam and lidar based sensor technologies, in: 2021 Mohammad Ali Jinnah University International Conference on Computing (MAJICC), IEEE, 2021, pp. 1--8.

\bibitem{macario2022comprehensive}
A.~Macario~Barros, M.~Michel, Y.~Moline, G.~Corre, F.~Carrel, A comprehensive survey of visual slam algorithms, Robotics 11~(1) (2022) 24.

\bibitem{li2020millimeter}
Y.~Li, Y.~Liu, Y.~Wang, Y.~Lin, W.~Shen, The millimeter-wave radar slam assisted by the rcs feature of the target and imu, Sensors 20~(18) (2020) 5421.

\bibitem{al2016ebsar}
S.~Al-Khalifa, M.~Al-Razgan, Ebsar: Indoor guidance for the visually impaired, Computers \& Electrical Engineering 54 (2016) 26--39.

\bibitem{lee2014wearable}
Y.~H. Lee, G.~Medioni, Wearable rgbd indoor navigation system for the blind, in: European Conference on Computer Vision, Springer, 2014, pp. 493--508.

\bibitem{altaha2016blindness}
I.~R. Altaha, J.~M. Rhee, Blindness support using a 3d sound system based on a proximity sensor, in: 2016 IEEE International Conference on Consumer Electronics (ICCE), IEEE, 2016, pp. 51--54.

\bibitem{slade2021multimodal}
P.~Slade, A.~Tambe, M.~J. Kochenderfer, Multimodal sensing and intuitive steering assistance improve navigation and mobility for people with impaired vision, Science robotics 6~(59) (2021) eabg6594.

\bibitem{sethi2022comprehensive}
D.~Sethi, S.~Bharti, C.~Prakash, A comprehensive survey on gait analysis: History, parameters, approaches, pose estimation, and future work, Artificial Intelligence in Medicine 129 (2022) 102314.

\bibitem{bensoussan2008evaluation}
L.~Bensoussan, J.-M. Viton, N.~Barotsis, A.~Delarque, Evaluation of patients with gait abnormalities in physical and rehabilitation medicine settings, Journal of Rehabilitation Medicine 40~(7) (2008) 497--507.

\bibitem{whittle2014gait}
M.~W. Whittle, Gait analysis: an introduction, Butterworth-Heinemann, 2014.

\bibitem{kirtley2006clinical}
C.~Kirtley, Clinical gait analysis: theory and practice, Elsevier Health Sciences, 2006.

\bibitem{gage1995gait}
J.~R. Gage, P.~A. Deluca, T.~S. Renshaw, Gait analysis: principles and applications, JBJS 77~(10) (1995) 1607--1623.

\bibitem{coutts1999gait}
F.~Coutts, Gait analysis in the therapeutic environment, Manual therapy 4~(1) (1999) 2--10.

\bibitem{jacquelin2010gait}
M.~Jacquelin~Perry, Gait analysis: normal and pathological function, New Jersey: SLACK (2010).

\bibitem{qiu2019body}
S.~Qiu, L.~Liu, Z.~Wang, S.~Li, H.~Zhao, J.~Wang, J.~Li, K.~Tang, Body sensor network-based gait quality assessment for clinical decision-support via multi-sensor fusion, Ieee Access 7 (2019) 59884--59894.

\bibitem{shih2012stable}
C.-L. Shih, J.~Grizzle, C.~Chevallereau, From stable walking to steering of a 3d bipedal robot with passive point feet, Robotica 30~(7) (2012) 1119--1130.

\bibitem{kale2004identification}
A.~Kale, A.~Sundaresan, A.~Rajagopalan, N.~P. Cuntoor, A.~K. Roy-Chowdhury, V.~Kruger, R.~Chellappa, Identification of humans using gait, IEEE Transactions on image processing 13~(9) (2004) 1163--1173.

\bibitem{veneman2007design}
J.~F. Veneman, R.~Kruidhof, E.~E. Hekman, R.~Ekkelenkamp, E.~H. Van~Asseldonk, H.~Van Der~Kooij, Design and evaluation of the lopes exoskeleton robot for interactive gait rehabilitation, IEEE Transactions on neural systems and rehabilitation engineering 15~(3) (2007) 379--386.

\bibitem{tao2012gait}
W.~Tao, T.~Liu, R.~Zheng, H.~Feng, Gait analysis using wearable sensors, Sensors 12~(2) (2012) 2255--2283.

\bibitem{zeng2024realistic}
T.~Zeng, G.~Loza~Galindo, J.~Hu, P.~Valdastri, D.~Jones, Realistic surgical image dataset generation based on 3d gaussian splatting, in: International Conference on Medical Image Computing and Computer-Assisted Intervention, Springer, 2024, pp. 510--519.

\bibitem{zeng2024yoco}
T.~Zeng, D.~He, F.~Yan, M.~He, Yoco: You only calibrate once for accurate extrinsic parameter in lidar-camera systems, arXiv preprint arXiv:2407.18043 (2024).

\bibitem{drnach2018identifying}
L.~Drnach, I.~Essa, L.~H. Ting, Identifying gait phases from joint kinematics during walking with switched linear dynamical systems, in: 2018 7th IEEE International Conference on Biomedical Robotics and Biomechatronics (Biorob), IEEE, 2018, pp. 1181--1186.

\bibitem{liu2016gait}
D.-X. Liu, X.~Wu, W.~Du, C.~Wang, T.~Xu, Gait phase recognition for lower-limb exoskeleton with only joint angular sensors, Sensors 16~(10) (2016) 1579.

\bibitem{gurchiek2019open}
R.~D. Gurchiek, R.~H. Choquette, B.~D. Beynnon, J.~R. Slauterbeck, T.~W. Tourville, M.~J. Toth, R.~S. McGinnis, Open-source remote gait analysis: A post-surgery patient monitoring application, Scientific reports 9~(1) (2019) 17966.

\bibitem{ozaki2011increases}
H.~Ozaki, M.~Sakamaki, T.~Yasuda, S.~Fujita, R.~Ogasawara, M.~Sugaya, T.~Nakajima, T.~Abe, Increases in thigh muscle volume and strength by walk training with leg blood flow reduction in older participants, Journals of Gerontology Series A: Biomedical Sciences and Medical Sciences 66~(3) (2011) 257--263.

\bibitem{lovejoy2005natural}
C.~O. Lovejoy, The natural history of human gait and posture: Part 2. hip and thigh, Gait \& posture 21~(1) (2005) 113--124.

\bibitem{courtine2003humanII}
G.~Courtine, M.~Schieppati, Human walking along a curved path. ii. gait features and emg patterns, European Journal of Neuroscience 18~(1) (2003) 191--205.

\bibitem{courtine2003humanI}
G.~Courtine, M.~Schieppati, Human walking along a curved path. i. body trajectory, segment orientation and the effect of vision, European Journal of Neuroscience 18~(1) (2003) 177--190.

\bibitem{patla1999online}
A.~E. Patla, A.~Adkin, T.~Ballard, Online steering: coordination and control of body center of mass, head and body reorientation, Experimental brain research 129 (1999) 629--634.

\end{thebibliography}




%
%
%
%
%
%
%
%
%
%
%

\end{document}